\documentclass[letterpaper]{article} 
\usepackage{aaai2027}  
\usepackage[hyphens]{url}  
\usepackage{graphicx} 
\usepackage{natbib}  
\usepackage{caption} 
\usepackage{algorithm}
\usepackage{algorithmic}
\usepackage{graphicx}
\usepackage{tikz}
\usepackage{multirow}
\usepackage{color}
\usepackage{makecell}
\usepackage{bbding}
\usepackage{booktabs, tabularx}
\usepackage{xcolor}
\usepackage{xpatch}
\usepackage{newfloat}
\usepackage{listings}
\usepackage{amsmath}
\usepackage{array} 
\usepackage{colortbl}
\usepackage{amssymb}
\definecolor{mygray}{gray}{.9}

\definecolor{clblue}{RGB}{222, 246, 246}

\definecolor{clblue}{RGB}{209, 246, 246}

\definecolor{clorange}{RGB}{255, 136, 16}

\definecolor{cgray}{RGB}{191, 191, 191}

\definecolor{lightred}{rgb}{1, 0.60, 0.63}
\definecolor{lightblue}{rgb}{0.55, 0.63, 1.0}
\definecolor{cvprblue}{rgb}{0.21,0.49,0.74}

\newcommand{\Rmnum}[1]{\uppercase\expandafter{\romannumeral #1}}
\usepackage{newfloat}
\usepackage{listings}
\DeclareCaptionStyle{ruled}{labelfont=normalfont,labelsep=colon,strut=off} 
\floatstyle{ruled}
\newfloat{listing}{tb}{lst}{}
\floatname{listing}{Listing}

\usepackage{booktabs}

\title{Towards UAV Image Dehazing: A UAV Atmospheric Scattering Model, Benchmark, and Geometry-Aware Deep Unfolding Network}
\author{
    Wenxuan Fang\textsuperscript{\rm 1}, 
    Jiangwei Weng\textsuperscript{\rm 1},
    Yu Zheng\textsuperscript{\rm 1},
    Junkai Fan\textsuperscript{\rm 1},
    Guangfa Wang\textsuperscript{\rm 1},\\
    Xiang Chen\textsuperscript{\rm 1},
    Jian Yang\textsuperscript{\rm 1},
    Jun Li\textsuperscript{\rm 1}*
}
\affiliations{
    \textsuperscript{\rm 1}School of Computer Science and Engineering, Nanjing University of Science and Technology

}

\begin{document}

\maketitle

\begin{abstract}
In UAV applications, haze significantly obscures distant details and weaken structural information, hindering the recovery of details. Current UAV scenarios still face two key challenges: (i) paired hazy/clean images from the real world are unobtainable, while the classical atmospheric scattering model is inadequate for modeling the spatially non-uniform haze in UAV imagery; (ii) existing dehazing methods struggle to remove the heavy haze accumulated in the upper regions of UAV images. To address these issues, we first propose a UAV Atmospheric Scattering Model (UASM), which explicitly incorporates flight altitude, viewing pitch, and extinction to characterize the non-uniform haze distribution in UAV imaging. 
Based on UASM, we develop a physics-driven dehazing framework, termed Geometry-aware Proximal Deep Unfolding Network (GP-DUN). Specifically, GP-DUN consists of three key modules: a Latent Geometry Estimator (LGE) that infers transmittance consistent with UAV imaging geometry, a Geometry-aware Gradient Descent Module (GeoGDM) that embeds UASM into the data-fidelity term and performs physics-consistent closed-form updates, and an Pooling-Expert Proximal Mapping Module (PE-PMM) that learns an implicit prior to restore textures and structures beyond the capability of explicit physical modeling. 
In addition, we further construct UASM-HazeSet, which provides controllable paired synthetic data together with 2,285 real UAV haze images for testing. Extensive experiments show that GP-DUN consistently outperforms existing methods on both UASM-HazeSet and real UAV haze benchmarks.
\end{abstract}


\section{Introduction}
\label{intro}

\begin{figure}[!t]
        \centering
        \includegraphics[width=0.95\linewidth]{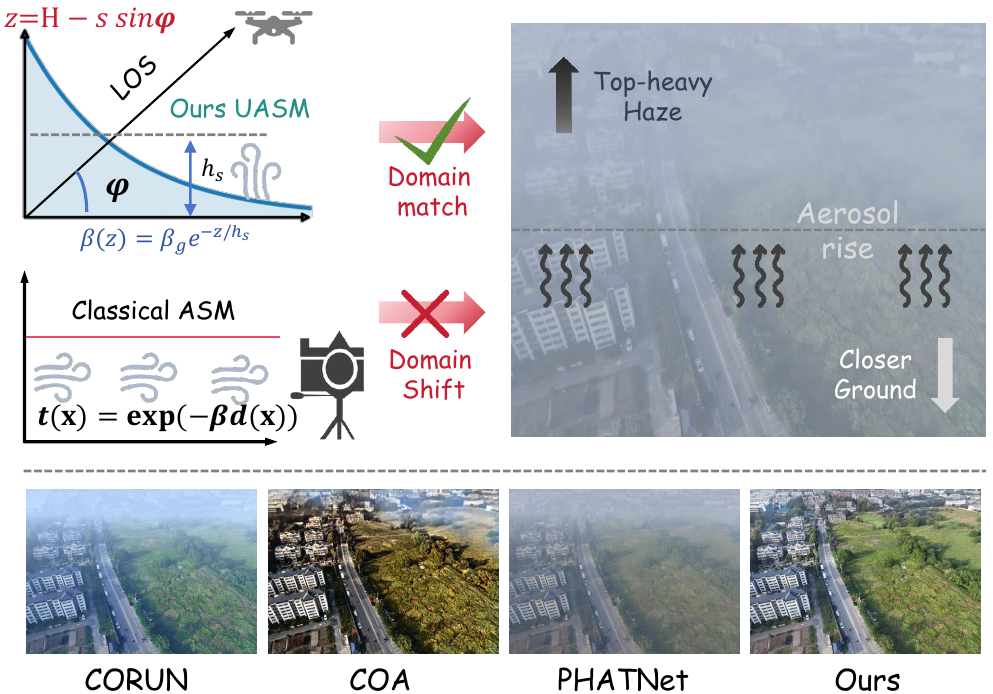}
        \vskip -0.1in 
        \caption{Motivation. Our UASM better synthesis top-heavy UAV haze than the conventional ASM, reducing the synthetic-to-real domain gap. Existing dehazing~\cite{corun,coa,phatnet} methods leave residual haze in the near-air region, our method produces clearer and more natural dehazing results.}
        \label{fig:motivation}
\end{figure}

\begin{figure*}[!t]
    \begin{center}
        \includegraphics[width=0.97\linewidth]{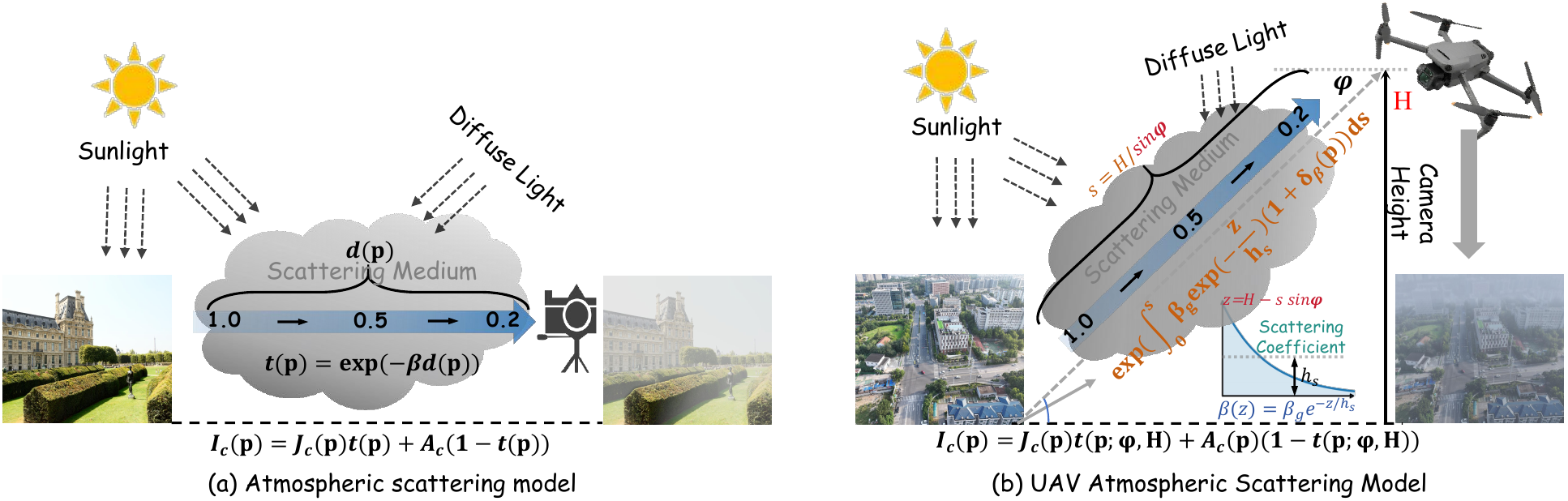}
    \end{center}
    \vskip -0.1in 
    \caption{Illustration of the classical atmospheric scattering model and the proposed UAV atmospheric scattering model (UASM). The classical ASM models haze formation with a depth-dependent transmittance under a homogeneous medium assumption, while UASM additionally considers flight altitude, viewing pitch, and altitude-varying extinction to describe the top-heavy non-uniform haze in UAV imaging.} 
    \label{fig:UASM}
\end{figure*}

UAVs have become an important visual sensing platform for urban inspection, traffic monitoring, and aerial mapping~\cite{ zhao2025hdcfn}. However, under hazy conditions, light scattering and absorption significantly reduce image contrast and wash out distant details, which directly degrades the reliability of downstream tasks such as object detection~\cite{fang2023multi}, semantic understanding~\cite{zhou2019semantic}. Therefore, developing robust dehazing methods tailored to UAV imagery is of both practical importance and research interest.

Despite existing dehazing methods~\cite{c2pnet, sgdn,fu2025iterative} have made significant progress in horizontal-view ground scenes, directly applying this methods to drone scenarios still faces challenges.
Mostly methods~\cite{wu2023ridcp,sgdn,fu2025iterative} implicitly assume spatially uniform haze in the image plane, resulting in unstable restoration for distant regions. 
However, UAV imaging at high altitudes and large pitch angles causes image pixels to correspond to light rays with significantly different viewing directions and path lengths, which typically leads to a characteristic top-heavy haze distribution~\cite{heald2011exploring}. 
As illustrated in Fig.~\ref{fig:motivation}, representative models (e.g., CORUN~\cite{corun}, COA~\cite{coa}, and PHATNet~\cite{phatnet}) still leave noticeable residual haze around the skyline. More importantly, to the best of our knowledge, there is still no dedicated benchmark for UAV image dehazing. However, in reality, it is almost impossible to obtain paired hazy/clean UAV images captured from the same viewpoint and under the same scene conditions. Most existing dehazing datasets~\cite{ancuti2020nh,li2018benchmarking} are acquired by ground-based cameras at low altitudes and near-horizontal viewing angles using atmospheric scattering models (ASM)~\cite{asm}; their degradation patterns differ fundamentally from those found in high-altitude UAV images.

As shown in Fig.~\ref{fig:UASM}(a), ASM assumes a homogeneous medium with a constant extinction coefficient and models the transmittance as $t(\mathbf{p})=\exp(-\beta d(\mathbf{p}))$.
However, in UAV imaging, haze formation is jointly influenced by flight altitude, viewing pitch angle, and vertical atmospheric stratification.
As a result, the standard ASM deviates noticeably from real UAV haze and is insufficient to capture its strong spatial variation.
To address this limitation, we propose a \emph{UAV Atmospheric Scattering Model} (UASM). 
As illustrated in Fig.~\ref{fig:UASM}(b), UASM explicitly accounts for flight altitude, viewing pitch angle, and altitude-dependent extinction, and models optical thickness via light-of-sight (LOS) integration, thereby better characterizing the geometry-dependent and spatially non-uniform haze formation in UAV imagery. 
Built upon UASM, we further construct the first dedicated UAV dehazing benchmark, termed \textbf{UASM-HazeSet}. 
By explicitly controlling pitch angle, flight altitude, and visibility, we synthesize 22,208 paired UAV images. 
In addition, we collect 2,285 real-world hazy UAV images captured under diverse altitudes and viewpoints to form a test set for evaluating the generalization and robustness of dehazing methods in practical UAV scenarios.


To effectively address haze restoration under large-pitch UAV imaging, we propose a \textbf{Geometry-aware Proximal Deep Unfolding Network (GP-DUN)} for UAV image dehazing. 
GP-DUN formulates the UASM-constrained dehazing process as an energy minimization problem and unrolls the corresponding optimization into an end-to-end network, where physically consistent updates and learned priors are alternately performed through deep unfolding.
Specifically, GP-DUN consists of three key modules: (1) a \textbf{Latent Geometry Estimator (LGE)} that explicitly injects UAV geometry and predicts pixel-wise transmittance to capture the top-heavy non-uniform degradation; (2) a \textbf{Geometry-aware Gradient Descent Module (GeoGDM)} that derives a closed-form geometry-aware update for progressive haze correction, enabling stable and physically meaningful restoration in distant regions with severe haze accumulation; and (3) an \textbf{Pooling-Expert Proximal Mapping Module (PE-PMM)} that learns an implicit frequency-aware prior via multi-expert fusion to recover textures and structures beyond the capacity of explicit physical modeling. 
Extensive experiments demonstrate that GP-DUN achieves clear advantages on \textbf{UASM-HazeSet} and delivers more stable visual quality with stronger cross-condition generalization on real UAV hazy images. Our main contributions as follows:

\begin{itemize}
    \item  To the best of our knowledge, we are the first to adapt the classical ASM to UAV viewpoints, introducing the UAV Atmospheric Scattering Model (UASM). Through explicit modeling of flight altitude, viewing pitch angle, and altitude-dependent extinction, UASM effectively characterizes the spatially non-uniform haze unique to UAV imagery.
    \item  We construct the UAV-oriented dehazing benchmark, \textbf{UASM-HazeSet}, which consists of a controllable synthetic paired subset (\textbf{UASM-S}, 22,208 images) and a real-world UAV haze test set (\textbf{UASM-R}, 2,285 images), enabling systematic evaluation under varying pitch angles, flight altitudes, and visibility conditions.
    \item  We propose GP-DUN, a geometry-aware proximal deep unfolding framework that integrates UASM-consistent closed-form gradient updates (GeoGDM) with a learned multi-expert proximal mapping prior (PE-PMM), achieving superior performance on UASM-HazeSet and improved robustness on real UAV hazy images.
\end{itemize}

\section{Related Works}
\textbf{Image Dehazing.}
Image dehazing aims to recover clear images from haze-degraded observations. Early methods mainly relied on physical priors, such as the dark channel prior~\cite{dcp} and color-lines~\cite{fattal2014dehazing}. With the rapid development of deep learning, substantial progress has been achieved. Some studies incorporate ASM-related priors~\cite{wang2025dehaze,fan2025non,deng2020hardgan} to improve restoration quality. For instance, Fan \textit{et al.}~\cite{fan2025non} estimate the transmission map for haze removal, while Deng \textit{et al.}~\cite{deng2020hardgan} reformulate ASM and introduce haze-aware representation distillation. Other works~\cite{dehazeformer,sgdn,guo2022image,cui2023focal,yang2024robust} directly learn the mapping from hazy to clean images, such as the Transformer-based method with transmission-aware positional embeddings proposed by Guo \textit{et al.}~\cite{guo2022image}. Beyond network architecture design, contrastive learning~\cite{fang2026weathercycle,c2pnet}, domain adaptation~\cite{phatnet,yang2024unified,zhang2025beyond}, and diffusion flow~\cite{shin2025hazeflow,lan2025schrodinger} have also been explored. For example, Cui \textit{et al.}~\cite{cui2023focal} enhance important spatial and frequency responses via a dual-domain selection mechanism, while HazeFlow~\cite{shin2025hazeflow} reformulates ASM as an ordinary differential equation. Despite these advances, existing methods still struggle to generalize to real-world UAV cross-domain haze, limiting their practical applicability.

\textbf{Deep Unfolding Image Restoration.} 
Deep unfolding networks (DUNs) bridge model-based optimization and data-driven learning, and therefore offer improved interpretability and flexibility compared with purely learning-based architectures. Recently, DUNs have been successfully applied to a wide range of image restoration~\cite{zhang2025joint, li2023pixel, yang2018proximal,he2025unfoldir,corun}. Among them, DGUN~\cite{mou2022deep} introduced a generic proximal-gradient formulation to learn image degradations. However, it does not explicitly disentangle prior knowledge from the degradation process, and relies on a single-path unfolding strategy to model complex degradations. CORUN~\cite{corun} further improves the dehazing framework by performing dual proximal gradient descent over the atmospheric scattering model and scene features. UnfoldIR~\cite{he2025unfoldir} introduced IDIR model with dedicated regularization terms for smoothing illumination and enhancing texture.
In contrast to previous unfolding-based dehazing methods, our approach explicitly couple geometry-aware physical modeling and a learnable proximal unfolding framework to handle top-heavy and spatially non-uniform haze in the UAV scences.

\section{UAV Atmospheric Scattering Model (UASM)}
\label{sec:uasm}
\textbf{Definition of Atmospheric Scattering Model.}
As shown in Fig.~\ref{fig:UASM}(a), given a hazy observation $\mathbf{I}\in\mathbb{R}^{H\times W\times 3}$, the ASM formulates the image formation as:
\begin{equation}
\mathbf{I}_c(\mathbf{p})=\mathbf{J}_c(\mathbf{p})\, t(\mathbf{p})+\mathbf{A}_c\left(1-t(\mathbf{p})\right), \quad c\in\{r,g,b\},
\label{eq:asm}
\end{equation}
where $\mathbf{p}$ denotes the pixel location, $\mathbf{J}$ is the haze-free scene, $\mathbf{A}\in\mathbb{R}^{3}$ is the atmospheric light, and $t(\mathbf{p})\in(0,1]$ is the transmittance. Under the common homogeneous-medium assumption, the extinction coefficient $\beta$ is treated as a constant, leading to $t(\mathbf{p})=\exp\!\left(-\beta\, d(\mathbf{p})\right),$ where $d(\mathbf{p})$ represents the distance associated with pixel $\mathbf{p}$.

\noindent\textbf{Limitations of ASM.}
For UAV imaging, flight altitude and camera pitch angle jointly determine the LOS length and the portion of atmosphere traversed by each ray, while aerosol density also varies with altitude. As a result, UAV haze usually exhibits a characteristic \emph{top-heavy} pattern~\cite{heald2011exploring}, with denser haze in the upper image regions and lighter haze near the bottom. 
This is because upper-image rays are generally more oblique and travel through a longer effective atmospheric path, leading to stronger scattering and lower transmittance. 
Therefore, the conventional ASM is inadequate for modeling UAV-specific haze formation. To this end, we explicitly incorporating UAV geometry and altitude-varying extinction, resulting in a UASM.


\begin{figure*}[!t]
    \begin{center}
        \includegraphics[width=0.96\textwidth]{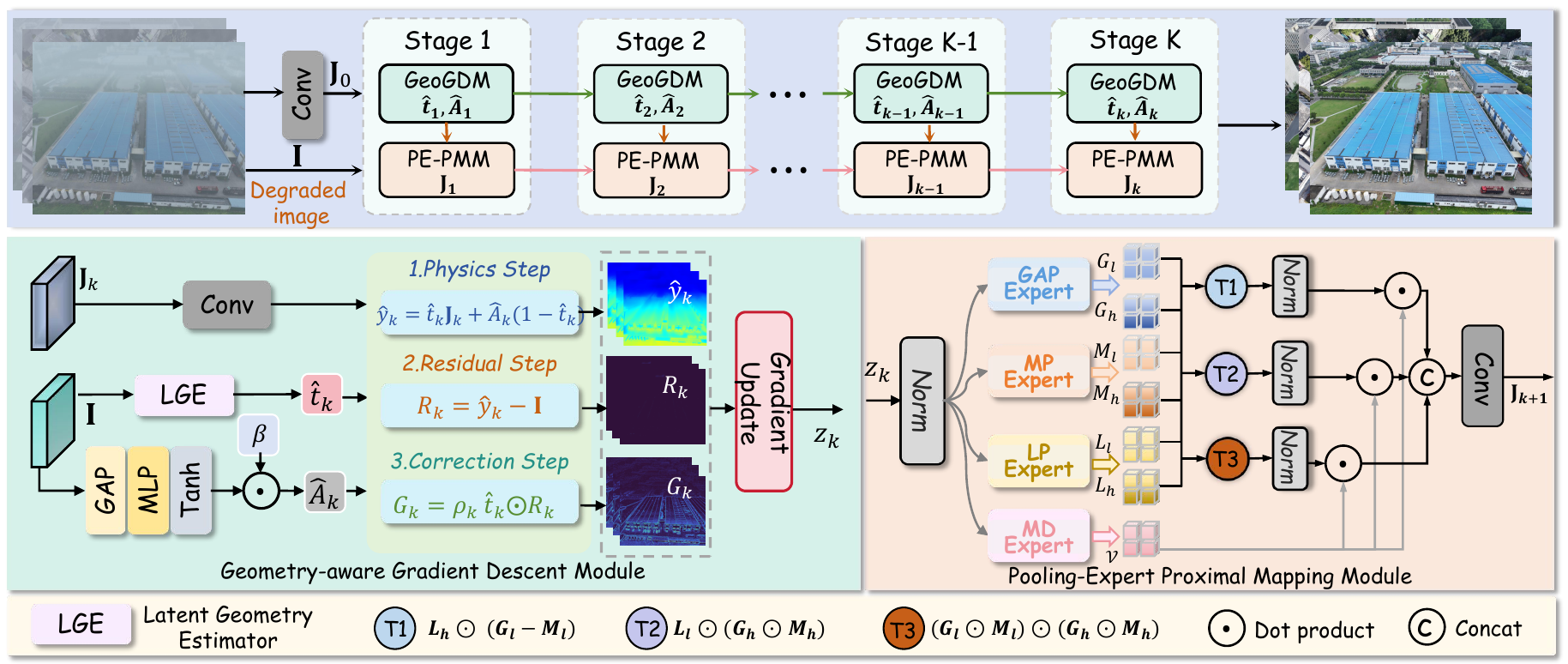}
    \end{center}
    \vskip -0.1in
    \caption{Architecture of the proposed GP-DUN. The network unrolls UAV image dehazing into multiple stages, where LGE estimates the geometry-aware transmittance, GeoGDM performs UASM-guided gradient updates, and PE-PMM approximates the proximal mapping with multi-expert fusion for texture and structure restoration.} 
    \label{fig:methods}
\end{figure*}


\noindent\textbf{Geometry and pitch-angle definition.}
As shown in Fig.~\ref{fig:UASM}(b), we define the ground plane as $\Pi: z=0$ and place the UAV camera center at altitude $H$.
For each pixel location $\mathbf{p}$, its LOS is parameterized by the traveled distance $s$, with $s=0$ corresponding to the camera center.
Following a UAV-oriented convention, the pitch angle $\varphi$ is defined as the angle between the LOS direction and the ground plane, such that the nadir view corresponds to $\varphi=90^\circ$.
Under this definition, the altitude of a point along the LOS is:
\begin{equation}
z(s)=H-s\sin\varphi, \qquad s\in[0,d(\mathbf{p})].
\label{eq:z_of_s_main}
\end{equation}

\noindent\textbf{Line-of-sight optical thickness.}
Unlike the classical ASM, which assumes a homogeneous medium with constant extinction, UAV imaging often occurs in a vertically stratified atmosphere, where the extinction decreases with altitude.
By further considering a slowly varying spatial perturbation, the optical thickness along the LOS can be written as:
\begin{equation}
\tau(\mathbf{p};\varphi,H)
=
\int_{0}^{d(\mathbf{p})}
\beta_g
\exp\!\left(
-\frac{H-s\sin\varphi}{h_s}
\right)
\Bigl(1+\delta_{\beta}(\mathbf{p})\Bigr)\,ds,
\label{eq:tau_main}
\end{equation}
where $\beta_g$ is the near-ground extinction coefficient, $h_s$ is the scale height controlling the vertical decay, and $\delta_{\beta}(\mathbf{p})$ denotes a slowly varying spatial perturbation satisfying $|\delta_{\beta}(\mathbf{p})|\ll 1$. $\beta_g$ can be linked to visibility $V$ via Koschmieder's law~\cite{lee2016visibility}, i.e., $\beta_g \approx 3.912/V$.

For ground-intersecting rays, where $d(\mathbf{p})=s=H/\sin\varphi$, the dominant geometry-aware component admits the following closed-form expression:
\begin{equation}
\tau(\mathbf{p};\varphi,H)
=
\frac{\beta_g h_s}{\sin\varphi}
\Bigl(1-e^{-H/h_s}\Bigr)
\Bigl(1+\delta_{\beta}(\mathbf{p})\Bigr).
\label{eq:tau_angle_main}
\end{equation}
Eq.~\eqref{eq:tau_angle_main} shows that the optical thickness increases as the pitch angle decreases, which explains the top-heavy haze pattern commonly observed in UAV imagery. The transmittance is defined as:
\begin{equation}
t(\mathbf{p};\varphi,H)=\exp\!\big(-\tau(\mathbf{p};\varphi,H)\big).
\label{eq:t_main}
\end{equation}

Based on the above geometry-aware transmittance, the UAV haze formation process is modeled as:
\begin{equation}
\mathbf{I}_c(\mathbf{p})
=
\mathbf{J}_c(\mathbf{p})\,t(\mathbf{p};\varphi,H)
+
\mathbf{A}_c(\mathbf{p})\Bigl(1-t(\mathbf{p};\varphi,H)\Bigr),
\ \ c\in\{r,g,b\},
\label{eq:uasm_main}
\end{equation}
\textbf{The complete derivation is provided in Appendix.2.}

\section{Methodology}

Based on UASM, we transform the UAV image dehazing problem into a  physics-driven energy minimization problem.
Given an observed hazy image $\textbf{I}$, our goal is to recover the latent clear image $\textbf{J}$ by minimizing:
\begin{equation}
\min_{\textbf{J}}\; \mathcal{L}(\textbf{J})
=
\underbrace{\frac{1}{2}\left\|\mathcal{D}_{uasm}(\textbf{J})-\textbf{I}\right\|_2^2}_{\text{Data fidelity}}
+\lambda\underbrace{g_{\theta}(\textbf{J})}_{\text{Implicit prior}},
\label{eq:energy}
\end{equation}
where $\mathcal{D}_{uasm}(\cdot)$ denotes the Eq.\ref{eq:uasm_main} and $g_{\theta}(\cdot)$ represents an implicit image prior learned by a neural network. We solve Eq.~\ref{eq:energy} via proximal gradient descent (PGD), which alternates between a physics-driven gradient step and a proximal mapping step. \textbf{The detailed solution process is in the Appendix.5.}

\noindent\textbf{1) Physics-driven gradient step.}
Let $\mathcal{D}(x)=\frac{1}{2}\|\mathcal{D}_{uasm}(\textbf{J})-\textbf{I}\|_2^2$. 
Since $\partial \mathcal{D}_{uasm}/\partial\textbf{J} = t$, the gradient admits a closed form:
\begin{equation}
\nabla_\textbf{J} \mathcal{D}(\textbf{J})
=
t\odot\big(\textbf{J}\odot t + A\odot(1-t)-\textbf{I}\big).
\label{eq:data_grad}
\end{equation}
The model alternately updates \textbf{J} during the iteration process. The following section will use the $k$-th iteration ($1 \leq k\leq K$) to illustrate the alternative solution process. Accordingly, we update an intermediate variable $z^{k}$ by
\begin{equation}
z^{k}=\textbf{J}^{k-1}-\rho_k\,\nabla_\textbf{J} \mathcal{D}(\textbf{J}^{k-1}),
\label{eq:gd_update}
\end{equation}
where $\rho_k$ denotes the step size. 
This step is implemented by our GeoGDM to perform geometry-consistent corrections.

\noindent\textbf{2) Proximal mapping step.}
Next, PGD solves the subproblem associated with the implicit prior:
\begin{equation}
\textbf{J}^{k}=\mathrm{prox}_{\rho_k\lambda g_{\theta}}(z^{k}).
\label{eq:prox_update}
\end{equation}


The PGD iterations in Eqs.~\eqref{eq:gd_update}-\eqref{eq:prox_update} are unrolled into a $K$-stage network, with the number of stages set to $K=7$.

\subsection{Geometry-aware Gradient Descent Module}
\label{sec:geogdm}

To explicitly capture UAV geometry-dependent scattering mechanism, we design the \textbf{Geometry-aware Gradient Descent Module (GeoGDM)} that transforms the overall physical process into learnable geometric inferences based on the closed-form solution in Eq.~\ref{eq:gd_update}.

As shown in Fig.~\ref{fig:methods}, given the stage-$k$ latent clean estimate $\textbf{J}_k$ and the hazy image $\textbf{I}$, GeoGDM first infers the physical factors required by the UASM model, then re-projects $\textbf{J}_k$ to the observation space to form a data-fidelity residual, and finally updates $\textbf{J}_k$ along the UASM consistent descent direction. We first estimate the transmittance map $\hat{t}_k$ via the proposed \textbf{Latent Geometry Estimator (LGE)}. 
By injecting normalized image-plane coordinates, LGE endows the network with spatial awareness and predicts a spatial distribution map $\mathbf{M}$ together with a global intensity scalar $\alpha$:
\begin{equation}
\hat{t}_{k}=\mathcal{H}_{LGE}(\mathbf{I})=\exp\!\left(-\alpha \odot \mathbf{M}\right),
\label{eq:lge_tk}
\end{equation}
where $\mathbf{M}\!\approx\!1/\sin\varphi$ serves as a proxy of the geometry-induced path variation, and $\alpha\!\approx\!\beta_g h_s(1-e^{-H/h_s})$ summarizes the scene-level haze strength (details in .~Alg.~\ref{alg_lge}). 

To robustly estimate the airlight, we aggregate global scene statistics through global average pooling and a MLP, followed by a Tanh nonlinearity and a learnable scale $\beta$:
\begin{equation}
\hat{A}_{k}=\beta \odot \tanh\!\Big(\mathcal{M}_{mlp}\big(\mathcal{G}(\textbf{I})\big)\Big),
\label{eq:A_hat_k}
\end{equation}
where $\mathcal{G}(\cdot)$ denotes GAP and $\mathcal{M}_{mlp}(\cdot)$ denotes the MLP.

With $(\hat{t}_k,\hat{A}_k)$, we re-project the current estimate back to the observation space using the UASM forward operator:
\begin{equation}
\hat{y}_k = \textbf{J}_k \odot \hat{t}_k + \hat{A}_k \odot (1-\hat{t}_k),
\quad
R_k=\hat{y}_k-\textbf{I},
\label{eq:reproj_residual}
\end{equation}
where $R_k$ measures the physical inconsistency between the rendered observation and the input, and it is exactly the error term of the UASM-constrained data-fidelity energy.

\begin{algorithm}[t]
\caption{Latent Geometry Estimator (LGE)}
\begin{algorithmic}[1]
\STATE \textbf{Input:} Hazy image $\textbf{I}\in\mathbb{R}^{B\times 3\times H\times W}$
\STATE \textbf{Output:} physical transmittance map $\mathbf{\hat{t}}_{k}\in(0,1]^{B\times 1\times H\times W}$

\STATE \textbf{Step 1: Geometry-aware coordinate injection}
\STATE $\mathbf{G}_x,\mathbf{G}_y \leftarrow \textsc{Meshgrid}([-1,1])$ \COMMENT{normalized coordinates}
\STATE $\mathbf{G}_{in} \leftarrow \textsc{Concat}(\textbf{I},\mathbf{G}_x,\mathbf{G}_y)$ \COMMENT{$B\times 5\times H\times W$}
\STATE $\mathbf{F} \leftarrow \sigma\!\left(\textsc{Conv}_{3\times 3}(\mathbf{G}_{in})\right)$ \COMMENT{$\sigma$: LeakyReLU}

\STATE \textbf{Step 2: Spatial distribution (proxy for $1/\sin\varphi$)}
\STATE $\mathbf{M} \leftarrow \textsc{Softplus}\!\big(\textsc{SpatialNet}(\mathbf{F})\big)$ \COMMENT{$\mathbf{M}(\mathbf{p})\ge 0$}

\STATE \textbf{Step 3: Global intensity (proxy for haze-strength)}
\STATE $\alpha \leftarrow \textsc{Softplus}\!\big(\textsc{MLP}(\textsc{GAP}(\mathbf{F}))\big)$ \COMMENT{$\alpha \ge 0$}

\STATE \textbf{Step 4: Physics-driven synthesis}
\STATE $\mathbf{\hat{t}}_{k} \leftarrow \exp\!\left(-\alpha \odot \mathbf{M}\right)$
\STATE \COMMENT{UASM alignment: $\tau(\mathbf{p})\approx \alpha\,\mathbf{M}(\mathbf{p})$, where $\mathbf{M}(\mathbf{p})\!\approx\!1/\sin\varphi$ and $\alpha\!\approx\!\beta_g h_s(1-e^{-H/h_s})$}
\STATE \textbf{return} $\mathbf{\hat{t}}_{k}$.
\end{algorithmic}
\label{alg_lge}
\end{algorithm}

\noindent\textbf{Closed-form geometry-aware gradient descent.}
Then, GeoGDM converts the physical residual into a mathematically consistent descent direction. Since the UASM forward model is linear with respect to $\textbf{J}_k$ and its element-wise Jacobian satisfies $\partial \hat{y}_k/\partial \textbf{J}_k=\hat{t}_k$, the gradient of the data-fidelity term reduces to a transmittance-weighted residual (cf.~Eq.~\ref{eq:data_grad}). 
Therefore, GeoGDM performs the stage-wise update:
\begin{equation}
z_{k}=\textbf{J}_k-\rho_k\,\hat{t}_k \odot R_k,
\label{eq:geogdm_update}
\end{equation}
where $\rho_k$ is a learnable step size. 
Intuitively, $\hat{t}_k$ acts as a geometry-aware sensitivity map: in heavy-haze regions (small $\hat{t}_k$), the observation is dominated by airlight and is less informative about $\textbf{J}_k$, hence the update is naturally attenuated to avoid unstable corrections.

\subsection{Pooling-Expert Proximal Mapping Module}
\label{sec:pepmm}

Relying solely on physics-driven gradient updates often fails to recover fine details. 
To overcome this limitation, we design the \textbf{Pooling-Expert Proximal Mapping Module (PE-PMM)} which consists of multiple proximal frequency domain experts to approximate the proximal operators in PGD and inject learnable implicit priors at each expansion stage.

Given the input feature $z_k\in\mathbb{R}^{B\times C\times H\times W}$ from the output of GeoGDM, we first obtain the normalized feature $\tilde{z}_k=\mathrm{Norm}(z_k)$. 
PE-PMM then employs a set of pooling experts $\{\mathcal{P}_m\}_{m=1}^{4}$ to extract complementary perspectives from $\tilde{z}_k$. 
Each expert decomposes the feature into a pooled low-frequency component and a high-frequency component, followed by learnable channel-wise reweighting:
\begin{equation}
\widetilde{L}_m=W_m^{l}\odot \mathcal{P}_m(\tilde{z}_k),\qquad
\widetilde{H}_m=(\mathbf{1}+W_m^{h})\odot\bigl(\tilde{z}_k-\mathcal{P}_m(\tilde{z}_k)\bigr),
\label{eq:unified_expert}
\end{equation}
where $W_m^{l},W_m^{h}$ are learnable channel-wise parameters, and $\odot$ denotes element-wise multiplication. 
The expert set includes global average pooling, local average pooling, and max pooling, and denote their outputs as $(G_L,G_H),(L_L,L_H), (M_L,M_H)$, which respectively capture global smooth context, local structural and salient responses. 
We further introduce a context modulation expert to provide a larger-receptive-field structural gate. 
Its high-frequency response is produced by a dilated max pooling operator:
\begin{equation}
H_d={E}_{\mathrm{DIL\text{-}MAX}}(\tilde{z}_k),
\qquad L_d=\tilde{z}_k-H_d,
\label{eq:dilate_split}
\end{equation}
and the context modulation output is defined as
\begin{equation}
V = W_l^{(d)}\odot L_d + (\mathbf{1}+W_h^{(d)})\odot H_d \;+\; a\odot(\tilde{z}_k\odot H_d)\;+\; b\odot \tilde{z}_k,
\label{eq:context_expert}
\end{equation}
where $W_l^{(d)},W_h^{(d)},a,b\in\mathbb{R}^{C\times 1\times 1}$ are learnable parameters. 
$V$ serves as a unified context gate to modulate subsequent interaction features, so that distant-haze regions rely more on large-scale structural consistency, while near-haze regions preserve more details.

\begin{figure*}[!t]
    \begin{center}
        \includegraphics[width=0.98\textwidth]{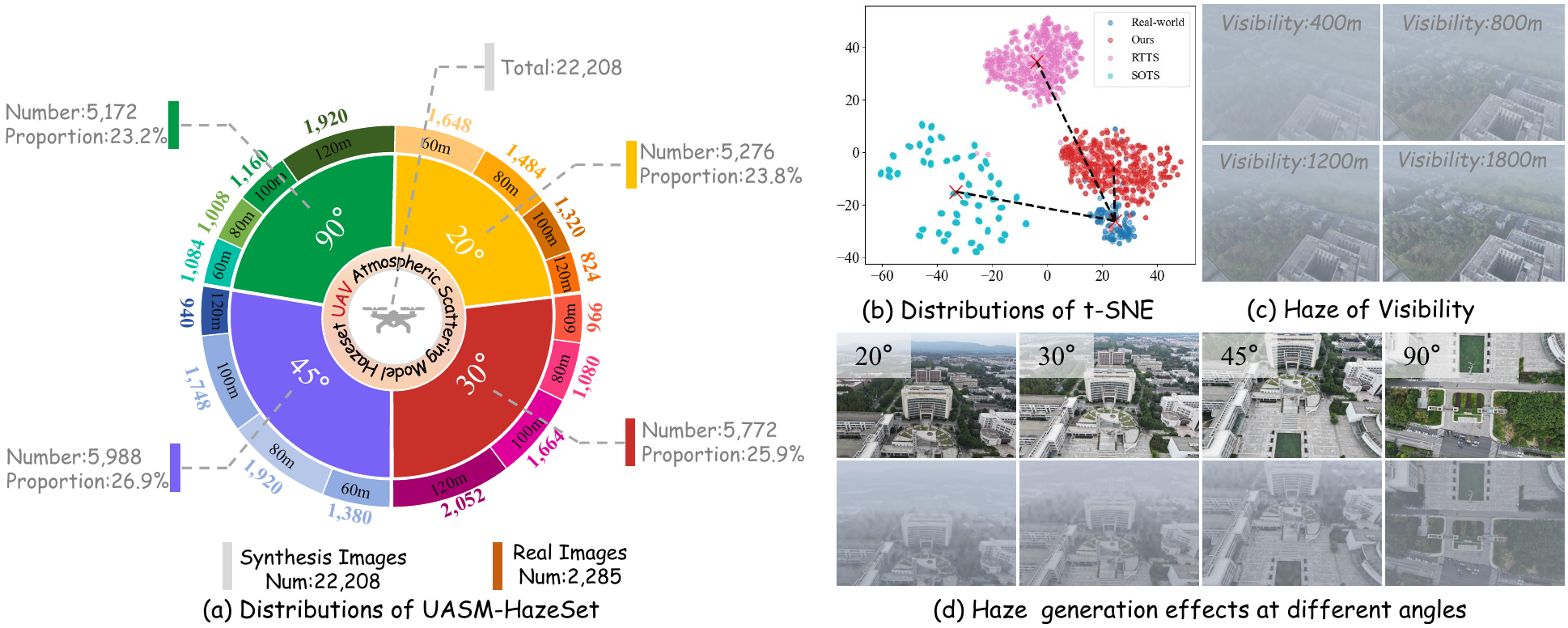}
    \end{center}
    \vskip -0.1in 
    \caption{(a) Dataset distributions. (b) t-SNE comparison of feature distributions. (c) Haze generation results under different visibility levels. (d) Synthesized UAV haze examples at different pitch angles.} 
    \label{fig:dataset}
\end{figure*}

After obtaining the low/high-frequency components, we construct three complementary interaction features to jointly model (i) low-frequency discrepancy, (ii) high-frequency agreement, and (iii) low-high coherence:
\begin{align}
T_1 &= \mathrm{Norm}\!\Big(L_h \odot (G_l - M_l)\Big), \label{eq:T1}\\
T_2 &= \mathrm{Norm}\!\Big(L_l \odot (G_h \odot M_h)\Big), \label{eq:T2}\\
T_3 &= \mathrm{Norm}\!\Big((G_h \odot M_h)\odot(G_l \odot M_l)\Big), \label{eq:T3}
\end{align}
where $T_1$ modulates local high-frequency responses by low-frequency discrepancy to suppress unreliable textures; $T_2$ strengthens structure responses that are consistent across experts; and $T_3$ emphasizes regions where low/high-frequency cues are simultaneously coherent. 
We then apply the context gate $V$ to all interaction features:
\begin{equation}
\tilde{T}_i = T_i \odot V,\quad i\in\{1,2,3\}.
\label{eq:gate}
\end{equation}

Finally, we concatenate the gated features and fuse them with a $1\times 1$ convolution to obtain a proximal correction term, followed by a residual output for stable refinement:
\begin{equation}
\Delta x_k=\mathrm{Conv}_{1\times 1}\big([\tilde{T}_1,\tilde{T}_2,\tilde{T}_3]\big),\qquad
x_{k+1}=z_k+\Delta x_k.
\label{eq:fuse_residual}
\end{equation}
This design enables PE-PMM to realize a learnable proximal mapping via frequency-complementary expert decomposition and interaction based fusion.

\noindent\textbf{Loss Functions.} Given clear target $x^{\ast}$, the network outputs a restored image $\hat{x}$. 
Our training objective is a weighted combination of pixel-wise fidelity, structural consistency, and frequency preservation:
\begin{equation}
\mathcal{L}
=
\lambda_1\,\mathcal{L}_{1}(\hat{x},x^{\ast})
+
\lambda_{s}\,\mathcal{L}_{\mathrm{SSIM}}(\hat{x},x^{\ast})
+
\lambda_{f}\,\mathcal{L}_{\mathrm{FFT}}(\hat{x},x^{\ast}).
\label{eq:loss_total}
\end{equation}
where $\ell_1$ is the L1 loss, $\mathcal{L}_{\mathrm{SSIM}}$ is the SSIM loss to preserve structural content. $\mathcal{L}_{\mathrm{FFT}}$ is the FFT loss to retain textures and suppress over-smoothing. $\lambda_1, \lambda_s, \lambda_f$ are the weights of each loss, which are 0.5, 0.3, and 0.1, respectively.

\begin{table*}[t]
\setlength{\tabcolsep}{0.4mm}
\caption{Quantitative comparison with recent state-of-the-art methods on UASM-S, RW$^2$AH, and UASM-R.}
\newcolumntype{?}{!{\vrule width 2pt}}
\newcolumntype{C}{>{\centering\arraybackslash}X}
\begin{center}
\begin{tabularx}{\textwidth}{l|c|*{4}{C}|*{4}{C}|*{3}{C}|*{1}{C}}
\toprule[0.15em]
\rowcolor{gray!20} & ~ & \multicolumn{4}{c|}{UASM-S} & \multicolumn{4}{c|}{RW$^2$AH} & \multicolumn{3}{c|}{UASM-R} & \\
\rowcolor{gray!20}
    \multirow{-2}{*}{Methods}  & \multirow{-2}{*}{Venus} & PSNR$\uparrow$ & SSIM$\uparrow$ & FADE$\downarrow$ & NIQE$\downarrow$ & PSNR$\uparrow$ & SSIM$\uparrow$ & FADE$\downarrow$ & NIQE$\downarrow$  & FADE$\downarrow$ & NIQE$\downarrow$ & NIMA$\uparrow$ & \multirow{-2}{*}{Params}  \\
\midrule[0.15em] 

DehazeFormer & TIP'23 & 18.63 & 0.631 & 1.036 & 4.623 & 20.36 & 0.612  & 0.690 & 5.806 & 1.310 & 3.106 & 3.647 & 25.44M  \\

C$^2$PNet& CVPR'23 & 17.64 & 0.639 & 0.912 & 4.614 & 15.51 & 0.472 & 0.615 & 7.012 & 1.314 & 3.415 & 3.938 & 7.17M  \\

CORUN & NIPS'24 & 19.85 & 0.664 & 1.261 & 4.224 & 16.68 & 0.550 & 0.533 & 5.399 & 0.403 & 3.052 & 4.423 & 18.67M \\

DCMPNet& CVPR'24 & 20.51 & 0.307 & 0.818 & 4.951 & 20.13 & 0.587 & 0.472 & 5.654  & 0.889 & 4.162 & 4.068 & 7.16M \\

DEANet& TIP'24 & 20.63 & 0.678 & 0.603 & \textcolor{blue}{3.689} & 21.14 &  0.564& 0.804 & 5.439 & 0.530 & 3.732 & 4.167 & 3.65M \\

SGDN & AAAI'25 & 20.62 & 0.672 & 0.586 & 4.598 & \textcolor{blue}{22.26} & \textcolor{blue}{0.668} & \textcolor{red}{0.400} & \textcolor{blue}{5.008}  & 0.776 & 3.120 & 4.330 & 13.32M \\

COA  & CVPR'25 & 15.88 & 0.539 & \textcolor{blue}{0.544} & 6.292 & 15.72 & 0.480 & 0.572& 6.761 & 0.633 & 3.837 & 3.446 & 12.63M  \\

PHATNet & ICCV'25  & {19.81} & {0.651} & 0.676  & 4.268 & 16.79 & 0.503 & 0.563 & 6.140  & 1.036 & 3.067 & 4.450 & 26.0M  \\

HOGformer & AAAI'26  & \textcolor{blue}{21.63} & \textcolor{blue}{0.689} & 0.707  & 5.064 & 18.02 & 0.547 & 0.703 & 5.567  & \textcolor{blue}{0.364} & \textcolor{blue}{2.617} & \textcolor{blue}{4.472} & 16.64M  
\\

BiLaLoRA & CVPR'26  & \textcolor{blue}{22.10} & \textcolor{blue}{0.694} & 0.563 & 4.672 & 19.72 & 0.604 & 0.491 & 5.336  & 0.409 & 3.168 & 4.215 & 3.76M  \\ \bottomrule

\textbf{Ours}  & -  & \textcolor{red}{23.63} & \textcolor{red}{0.707} & \textcolor{red}{0.536} & \textcolor{red}{3.553} & \textcolor{red}{22.37} & \textcolor{red}{0.680}  & \textcolor{blue}{0.413} & \textcolor{red}{4.813} & \textcolor{red}{0.328} & \textcolor{red}{2.452} & \textcolor{red}{4.910} & {18.87M}   \\
\bottomrule[0.15em]
\end{tabularx}
\scriptsize
\label{tab:main_results}
\end{center}
\end{table*}

\section{UAV Atmospheric Scattering Model Hazeset}
\label{sec:uaset}

\textbf{Motivation.} In practice, the UAV flight trajectory cannot be exactly repeated, and the scene content cannot be guaranteed to remain unchanged across captures. 
As a result, it is almost impossible to obtain paired hazy/clean UAV images under identical viewpoints and scene conditions. Meanwhile, existing hazy datasets rarely consider UAV-specific factors such as flight altitude and viewing pitch, making it difficult to systematically evaluate cross-view generalization. 
Motivated by the proposed UASM, we therefore construct \textbf{UASM-HazeSet}, a physics-guided benchmark with controllable UAV factors for both training and evaluation.

\noindent\textbf{Dataset Organization and Distribution.}
We capture haze-free UAV images using a \emph{DJI Mavic 3 Pro}. The data are collected on clear days with good illumination, and the scenes cover typical UAV viewpoints in urban/campus environments, including buildings, roads, and vegetation. Due to airspace regulations, the flight altitude is restricted to $H\in\{60,80,100,120\}\,\mathrm{m}$. Based on these clear reference images, we generate paired hazy observations using the UASM. For UAV viewing geometry, we consider representative pitch angles $\varphi\in\{20^\circ,30^\circ,45^\circ,90^\circ\}$. All images are resized to $1920\times1080$ to standardize training and evaluation. As shown in Fig.~\ref{fig:dataset}(a), UASM-HazeSet consists of two parts: a synthetic \emph{paired} UAV dataset (UASM-S) and a real-world UAV haze set (UASM-R). The synthetic paired dataset contains 22,208 images in total, with pitch-angle proportions of 23.8\% for $20^\circ$, 25.9\% for $30^\circ$, 26.9\% for $45^\circ$, and 23.2\% for $90^\circ$. Within each pitch-angle subset, we further cover multiple flight altitudes, and each clear image is augmented into four haze severity levels by controlling the visibility. In addition, we collected 2,285 real high-altitude haze drone images at arbitrary heights and angles as a test set to evaluate the model's performance in real-world scenarios.

In addition, Fig.~\ref{fig:dataset}(b) compares the feature distribution of our dataset with real hazy data. Compared with existing benchmarks (\textbf{See Appendix.3.1}), UASM-HazeSet is closer to the real distribution, which helps reduce the synthetic-to-real domain gap.

\begin{figure*}[!t]
    \begin{center}
        \includegraphics[width=0.94\textwidth]{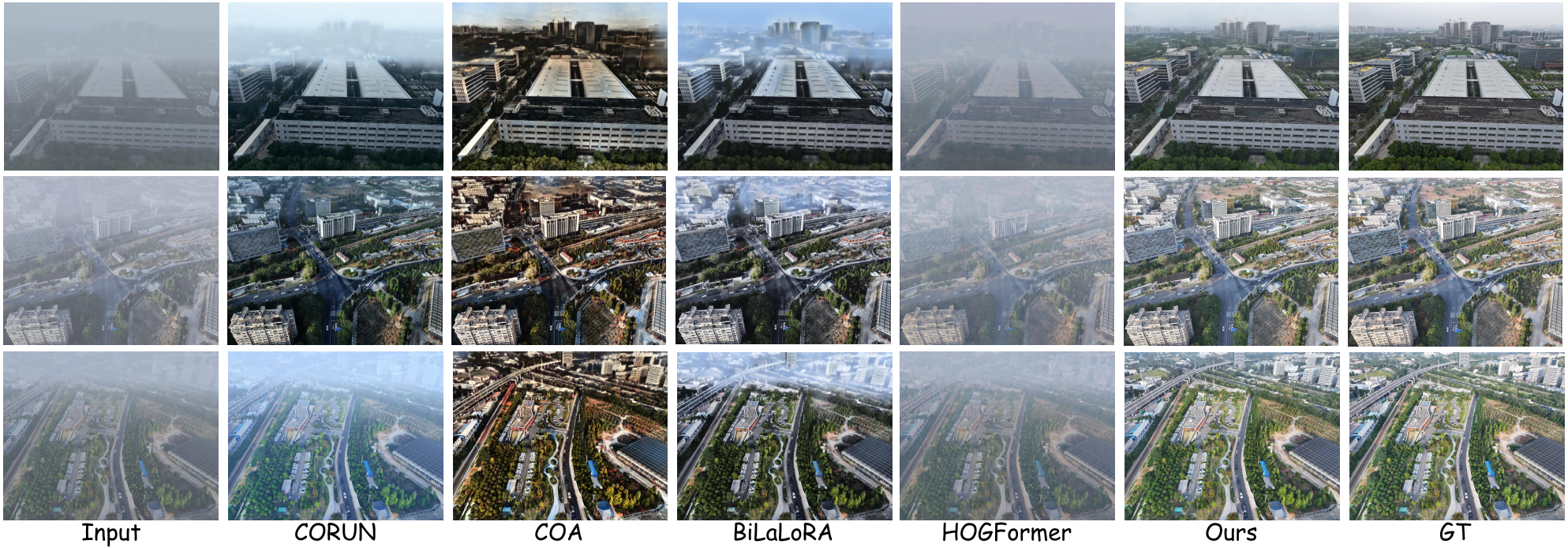}
    \end{center}
    \vskip -0.1in 
    \caption{Visual comparison on the UASM-S dataset. Our method removes spatially non-uniform haze more effectively and produces results that are visually closer to the ground truth.} 
    \label{fig:vis_uasms}
\end{figure*}

\begin{figure*}[!t]
    \begin{center}
        \includegraphics[width=0.94\linewidth]{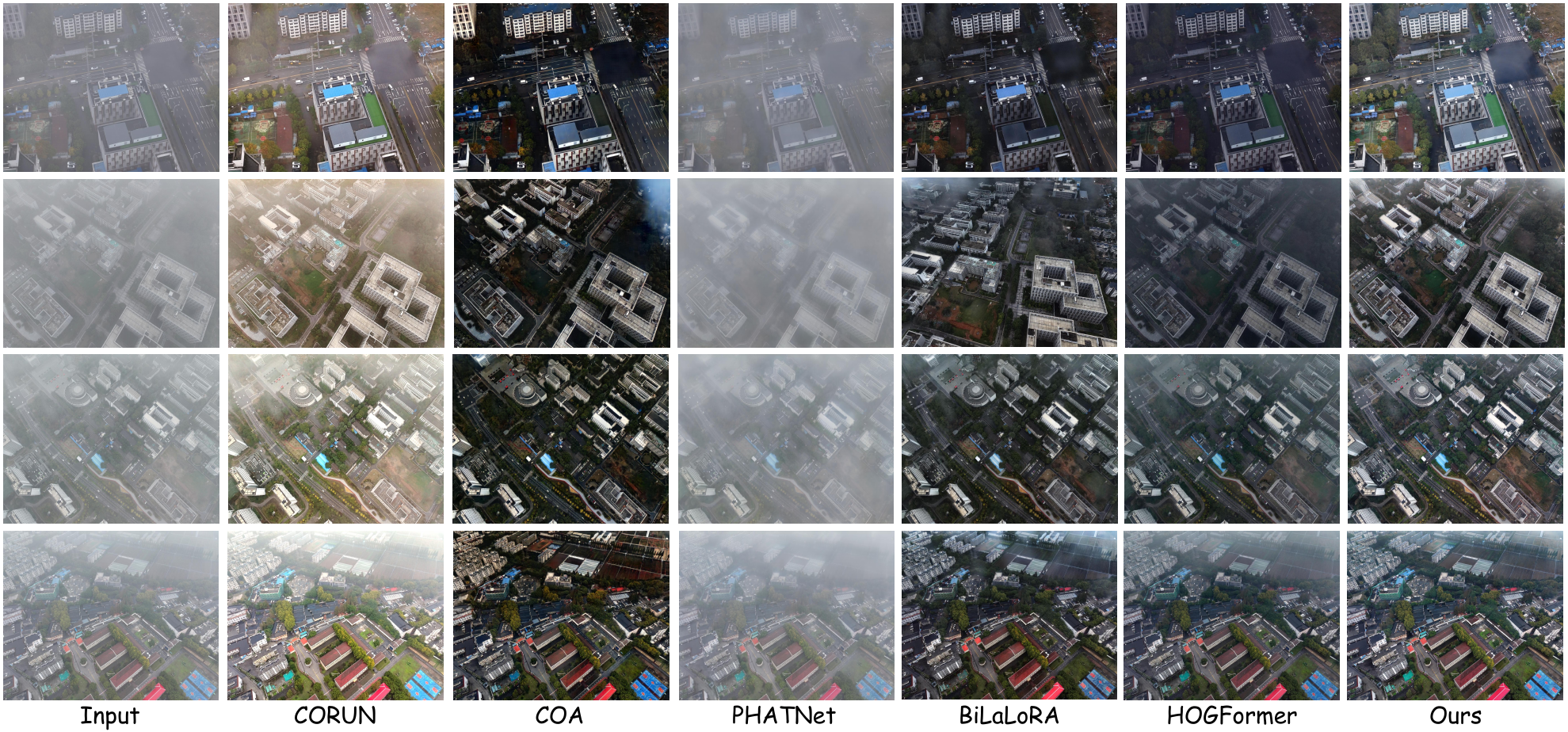}
    \end{center}
    \vskip -0.1in 
    \caption{Visualization results of representative dehazing effects on the UASM-R dataset were collected by DJI Mavic 3 Pro in a foggy urban environment. More real-world results and failure examples are in Appendix. 6 \& 7.} 
    \label{fig:real_uav}
\end{figure*}

\section{Experiments}

\label{sec:exp_settings}
\noindent\textbf{Datasets and evaluation metrics.}
We conduct experiments on \textbf{UASM-HazeSet} and \textbf{RW$^2$AH}~\cite{sgdn}. 
UASM-HazeSet includes 22,208 synthetic haze/clear UAV image pairs with varying pitch angles, flight altitudes, and visibility levels, together with a real-world UAV haze test set for evaluating practical generalization. 
RW$^2$AH~\cite{sgdn} contains 1,758 real-world well-aligned haze/clear pairs collected from stationary webcams. 
For paired evaluation, we use PSNR and SSIM; for real-world evaluation, we adopt FADE~\cite{choi2015referenceless}, NIQE~\cite{niqe}, and NIMA~\cite{talebi2018nima} to measure residual haze, perceptual naturalness, and overall image quality.

\subsection{Main Results}
\label{sec:main_results}

\begin{figure}[!t]
    \begin{center}
        \includegraphics[width=0.96\linewidth]{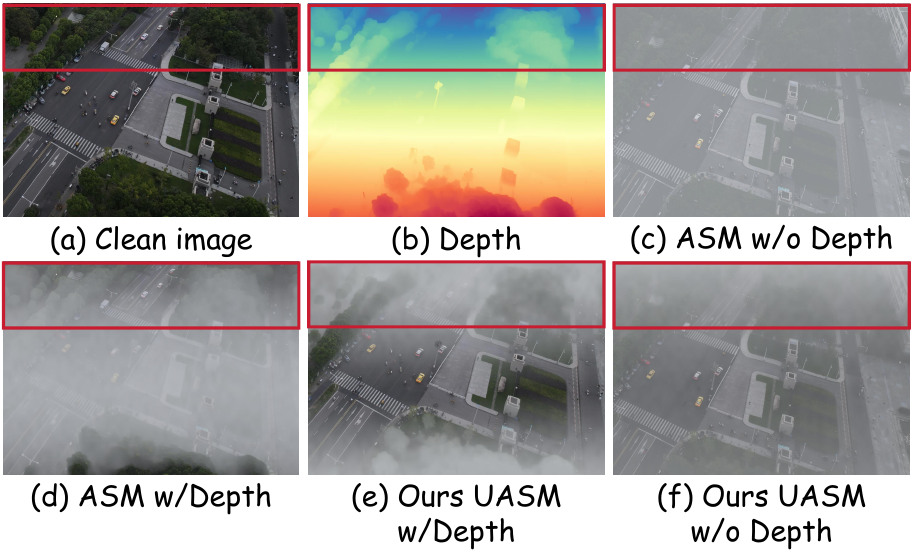}
    \end{center}
    \vskip -0.1in 
    \caption{Ablation on UASM. Compared with conventional ASM, depth-guided variants and our UASM.} 
    \label{fig:ablation_uasm}
\end{figure}

\begin{figure}[!t]
    \begin{center}
        \includegraphics[width=0.96\linewidth]{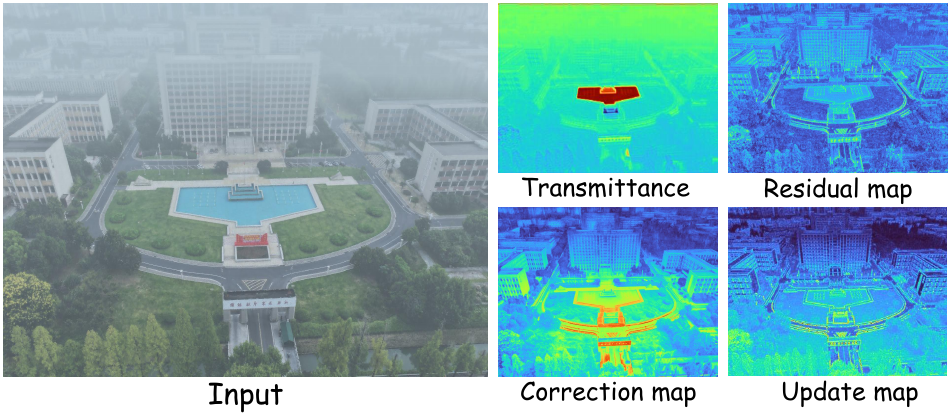}
    \end{center}
    \vskip -0.1in 
    \caption{Visualization of intermediate responses in GeoGDM, including the estimated transmittance map, residual map, correction map, and update map.} 
    \label{fig:feature_map}
\end{figure}

We compare GP-DUN with several recent state-of-the-art dehazing methods, including DehazeFormer~\cite{dehazeformer}, C$^2$PNet~\cite{c2pnet}, CORUN~\cite{corun}, DCMPNet~\cite{dcmpnet}, DEANet~\cite{deanet}, SGDN~\cite{sgdn}, COA~\cite{coa}, PHATNet~\cite{phatnet}, BiLaLoRA~\cite{bilalora} and HOGformer~\cite{hogformer}. For fair comparison, all competing methods are retrained under the same experimental protocol to achieve optimal performance. The quantitative results are reported in Table~\ref{tab:main_results}. \textbf{The results of RTTS is in Appendix 4.}

\noindent\textbf{Quantitative results.} On the proposed UASM-S benchmark, GP-DUN achieves the best performance across all reported metrics. Compared with the recently added baseline BiLaLoRA~\cite{bilalora}, GP-DUN improves PSNR \(+1.53\)dB and SSIM from 0.694 to 0.707, while further reducing FADE and NIQE to 0.536 and 3.553, respectively. Compared with the unfolding-based CORUN~\cite{corun}, GP-DUN also consistently yields better performance on all metrics. Compared with the strongest competitor, HOGFormer~\cite{hogformer}, GP-DUN improves PSNR by more than 9\% relatively and increases SSIM from 0.689 to 0.707, while also obtaining lower FADE and NIQE. 
These results indicate that our method not only improves distortion-based fidelity, but also produces clearer and more natural restorations. On RW$^2$AH, GP-DUN still delivers the best overall performance, despite this dataset not being UAV-specific. In particular, it surpasses SGDN~\cite{sgdn} and DEANet~\cite{deanet} in PSNR by 0.49\% and 5.82\%, respectively, and achieves the best SSIM and NIQE. Although its FADE is slightly inferior to SGDN, the overall results suggest that GP-DUN provides higher-quality restoration with better fidelity and perceptual quality. On the real-world UASM-R benchmark, GP-DUN shows even clearer advantages. It achieves the best FADE, NIQE, and NIMA, demonstrating stronger haze removal ability and better perceptual restoration on real UAV images.

\noindent\textbf{Qualitative results.} Fig.~\ref{fig:vis_uasms} presents qualitative comparisons on UASM-S. 
Existing methods, such as CORUN~\cite{corun}, COA~\cite{coa}, SGDN~\cite{sgdn}, BiLaLoRA~\cite{bilalora} and HOGFormer~\cite{hogformer}, often leave noticeable residual haze near the skyline and show unstable enhancement in the upper image regions. 
In contrast, GP-DUN removes spatially non-uniform haze more effectively and produces results that are visually closer to the ground truth. 
Fig.~\ref{fig:real_uav} further shows that, on real UAV haze images, our method achieves cleaner restoration with fewer contrast shifts and more faithful background colors than competing methods.

\begin{table}[t]
\centering
\vskip -0.1in 
\caption{Module ablation on UASM-S.}
\label{tab:ablation_module}
\setlength{\tabcolsep}{6pt}
\renewcommand{\arraystretch}{0.9}
\scalebox{0.8}{
\begin{tabular}{lcc}
\toprule
Method Variant & PSNR$\uparrow$ & SSIM$\uparrow$ \\
\midrule
w/o LGE & 23.22 & 0.671 \\
w/o GeoGDM & 20.17 & 0.663 \\
w/o PE-PMM & 19.49 & 0.626 \\
\rowcolor{red!8} Full GP-DUN & 23.63 & 0.707 \\
\midrule
\multicolumn{3}{c}{\textbf{PE-PMM design ablation}} \\
\midrule
Single expert & 22.34 & 0.676 \\
w/o context modulation & 23.56 & 0.684 \\
w/o T1 & 23.50 & 0.697 \\
w/o T2 & 23.55 & 0.684 \\
w/o T3 & 23.49 & 0.695 \\
\rowcolor{red!8} Full PE-PMM & 23.63 & 0.707 \\
\bottomrule
\end{tabular}}
\end{table}

\begin{table}[t]
\centering
\caption{Cross-dataset generalization results on UASM-R. All methods are trained on different source datasets and evaluated on UASM-R.}
\label{tab:cross_dataset_uasmr}
\setlength{\tabcolsep}{3.5pt}
\renewcommand{\arraystretch}{1}
\scalebox{0.82}{
\begin{tabular}{l|c|ccc}
\toprule
\multirow{2}{*}{Training Set} & \multirow{2}{*}{Method} & \multicolumn{3}{c}{UASM-R} \\
\cmidrule(lr){3-5}
& & FADE$\downarrow$ & NIQE$\downarrow$ & NIMA$\uparrow$ \\
\midrule
\multirow{4}{*}{RW$^2$AH}
& PHATNet    & 1.414 & 4.881 & 4.021 \\
& HOGFormer  & 0.698 & 3.704 & 4.267 \\
& BiLaLoRA   & 0.571 & 4.352 & 4.086 \\
\rowcolor{red!8}
& \textbf{Ours}       & \textbf{0.487} & \textbf{2.869} & \textbf{4.623} \\
\midrule
\multirow{4}{*}{SOTS-Outdoor}
& PHATNet    & 1.786 & 5.012 & 3.554 \\
& HOGFormer  & 0.843 & 4.221 & 3.196 \\
& BiLaLoRA   & 0.818 & 4.487 & 4.012 \\
\rowcolor{red!8}
& \textbf{Ours}       & \textbf{0.532} & \textbf{3.015} & \textbf{4.541} \\
\midrule
\multirow{4}{*}{UASM-S}
& PHATNet    & 1.036 & 3.067 & 4.450 \\
& HOGFormer  & 0.364 & 2.617 & 4.472 \\
& BiLaLoRA   & 0.409 & 3.168 & 4.215 \\
\rowcolor{red!8}
& \textbf{Ours}       & \textbf{0.328} & \textbf{2.452} & \textbf{4.910} \\
\bottomrule
\end{tabular}}
\end{table}

\subsection{Ablation Studies}
\noindent \textbf{Effect of UASM.}
We further compare different atmospheric scattering models for UAV haze synthesis in Fig.~\ref{fig:ablation_uasm}. 
The classical ASM assumes a homogeneous medium and thus produces nearly uniform haze over the image plane, which is inconsistent with the spatially varying degradation in UAV imagery. 
Introducing monocular depth estimated by DepthV2~\cite{yang2024depth} can partially increase spatial variation, but its performance is limited by inaccurate long-range depth prediction under aerial viewpoints. 
A similar limitation is observed when such estimated depth is directly incorporated into UASM. 
In contrast, our UASM explicitly models flight altitude, viewing pitch, and altitude-dependent extinction, enabling more realistic top-heavy and geometry-consistent haze synthesis without relying on unstable external depth estimation.

\noindent \textbf{Each Components.} Table~\ref{tab:ablation_module} presents the ablation results of the main components. Removing any major module leads to a consistent performance drop, which confirms that the proposed framework benefits from the complementarity among physics-guided optimization, and learned proximal refinement. In particular, the largest degradation is observed when PE-PMM is removed, indicating that the proximal prior plays a crucial role in recovering textures and structures beyond the capacity of the physical data term alone. We further verify the complementarity of the experts inside PE-PMM. Removing any of the interaction terms consistently degrades the performance, confirming that the proposed cross-frequency interactions are all effective for proximal refinement.
Fig.~\ref{fig:feature_map} illustrates the internal behavior of GeoGDM. 
Starting from the estimated transmittance map, the model first identifies the spatially varying haze distribution in the input image. 
Based on geometry-aware prior, the residual map highlights the regions where the current reconstruction is still inconsistent with the observation, especially in severely degraded areas. 
The subsequent correction map further concentrates on these haze-dominant regions, while the final update map filters out less relevant responses and preserves only the major restoration cues. 
Such a progressive refinement process verifies that GeoGDM effectively turns geometry-aware haze estimation into stable and targeted updates for UAV dehazing.

\noindent\textbf{Cross-dataset generalization on UASM-R.}
We further compare the training gap between other natural dehazing datasets and our proposed UASM-S, with the real UASM-R used for testing.
Table~\ref{tab:cross_dataset_uasmr} shows a clear trend: all methods perform substantially better when trained on \textbf{UASM-S} rather than RW$^2$AH or SOTS-Outdoor. 
This verifies that our UASM-S is better aligned with the degradation characteristics of UAV imagery. 
Among all methods, \textbf{Ours} consistently achieves the best results under all training settings, and its advantage becomes most evident when trained on UASM-S. 
These results confirm both the effectiveness of the proposed benchmark and the strong real-world generalization of GP-DUN in UAV dehazing.


\begin{figure}[!t]
    \begin{center}
        \includegraphics[width=0.96\linewidth]{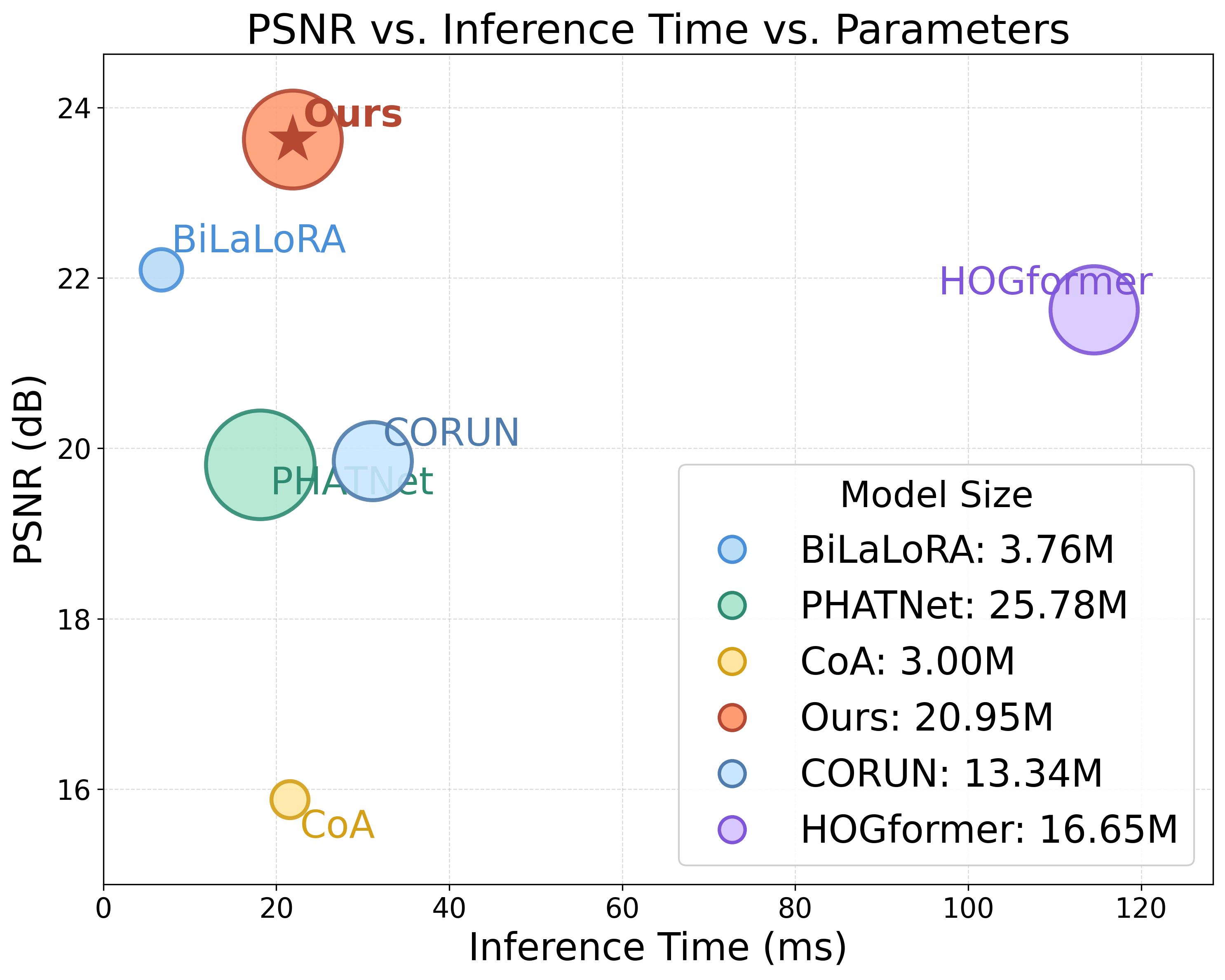}
    \end{center}
    \vskip -0.1in 
    \caption{Comparison of PSNR, inference time, and parameter count on $256\times256$ images. } 
    \label{fig:model_complexity}
\end{figure}


\noindent\textbf{Runtime comparison.}
Fig.~\ref{fig:model_complexity} compares the PSNR, inference time, and parameter count of different methods on $256\times256$ images, where the bubble size denotes the model size. 
Our method achieves the best PSNR while maintaining competitive inference efficiency. 
Specifically, although BiLaLoRA is slightly faster and lighter, its restoration performance is still inferior to ours. 
Compared with CORUN and PHATNet, our method achieves clearly higher PSNR with lower or comparable inference cost. 
In contrast, HOGformer requires substantially longer inference time while still delivering lower PSNR than our method. 
These results demonstrate that GP-DUN strikes a favorable trade-off among restoration quality, runtime efficiency, and model complexity.

\section{Conclusion}

In this paper, we introduce \textbf{UASM-HazeSet}, a physics-guided benchmark for UAV dehazing that explicitly models pitch, altitude, and visibility variations, and provides both synthetic paired data and real-world UAV haze images for evaluation. Based on this benchmark, we further propose \textbf{GP-DUN}, a geometry-aware deep unfolding framework that integrates UASM-consistent optimization with learned proximal refinement. Extensive experiments demonstrate that our method achieves superior performance on both UASM-HazeSet and real UAV hazy images. We hope that UASM-HazeSet will facilitate future research on UAV-oriented image restoration under adverse weather conditions.

\bibliography{aaai2027}


\end{document}